\pdfoutput=1

\documentclass[11pt]{article}

\usepackage[final]{acl}

\usepackage{times}
\usepackage{latexsym}

\usepackage[T1]{fontenc}

\usepackage[utf8]{inputenc}

\usepackage{microtype}

\usepackage{inconsolata}

\usepackage{graphicx}

\usepackage{times}
\usepackage{url}
\usepackage{latexsym}
\usepackage[utf8]{inputenc}
\usepackage{tcolorbox}
\usepackage{xcolor}
\usepackage{tikz-dependency}

\tcbset{
    colback=pink!5!white, 
    colframe=pink!75!black, 
    fonttitle=\bfseries, 
    boxrule=0.5mm, 
    arc=4mm, 
    outer arc=3mm,
    coltitle=black, 
    toptitle=1mm,
    bottomtitle=1mm
}

\title{Language Complexity Measurement as a Noisy Zero-Shot Proxy for
Evaluating LLM Performance}

\author{Birger Moell \\
 KTH Royal Institute of Technology\\
Stockholm, Sweden \\
 {\tt bmoell@kth.se} \\\And
 Johan Boye \\
  KTH Royal Institute of Technology\\
  Stockholm, Sweden \\
 {\tt jboye@kth.se} 
}

\begin{document}
\maketitle
\begin{abstract} Large Language Models (LLMs) have made significant strides in natural language generation but often face challenges in tasks requiring precise calculations and structural analysis. This paper investigates the performance of state-of-the-art LLMs on language complexity measurement tasks, through the computation of the LIX readability metric and Average Dependency Distance (ADD). Using Swedish high school and university-level essays, we evaluate the models' abilities to compute LIX scores and perform dependency parsing, comparing their results to established ground truths. Our findings reveal that while all models demonstrate some capacity for these tasks, ChatGPT-o1-mini performs most consistently, achieving the highest accuracy in both LIX computation and dependency parsing. Additionally, we observe a strong significant correlation -0.875  p 0.026 (N=6) between the models' accuracy in computing LIX and their overall performance on the Massive Multitask Language Understanding (MMLU) benchmark. These results suggest that language complexity measurement abilities can serve as a noisy zero-shot proxies for assessing the general capabilities of LLMs, providing a practical method for model evaluation without the need for extensive benchmarking datasets. \end{abstract}

\section{Introduction}
Large Language Models (LLMs) are advancing at an unprecedented pace. Until recently, models like ChatGPT demonstrated remarkable language generation capabilities but often struggled with mathematical and analytical tasks requiring a single correct answer \cite{wang2024can, li2024llms, mittal2024puzzlebench, wang2024re}.

This paper examines the performance of state-of-the-art LLMs on two analytical tasks related to language complexity: (1) computing the LIX readability metric and (2) performing dependency parsing to calculate the Average Dependency Distance metric. These tasks are particularly relevant as they test the mathematical capabilities of LLMs (in the case of LIX) and their structural reasoning abilities (in the case of dependency parsing). The LIX metric is especially intriguing because it involves counting the number of letters in tokens, a task that poses challenges given that tokens in LLMs are represented as numerical identifiers. This transformation should, theoretically, obscure information about the internal structure of words.

We evaluate six models: Gemini-1.5-Pro, Gemini-2.0-flash, (developed by Google), Llama 70b, Llama70b 3.3 (from Meta), GPT4o-mini, and o1-mini (released by OpenAI). Each model is presented with identical prompts for both tasks, and the outputs are compared against ground truth. The complete set of prompts is provided in the Appendix.


\section{Language complexity metrics}
\subsection{Average Dependency Distance}
{\em Average Dependency Distance (ADD)}, suggested by \citet{liu2008dependency}, calculates the distance between each word (except the root word) and its head in a dependency tree \citep{kubler2009dependency}. For example, in the dependency tree for the sentence ``John made the pie in the fridge'' below, the distance from ``fridge'' to ``pie'' is 3. The ADD of the sentence is 12/7, or about 1.7.
  
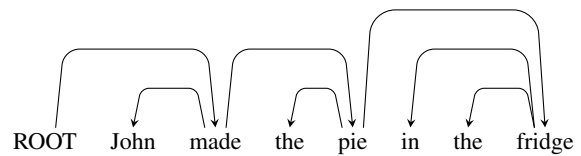
\begin{center}
    \begin{small}
    \begin{dependency}[hide label]
      \begin{deptext}[column sep=0.3cm]
        ROOT \& John \& made \& the \& pie \& in \& the \& fridge \\
      \end{deptext}
      \depedge{1}{3}{root}
      \depedge{3}{2}{nsubj}
      \depedge{3}{5}{obj}
      \depedge{5}{4}{det}
      \depedge{5}{8}{nmod}
      \depedge{8}{6}{case}
      \depedge{8}{7}{det}
    \end{dependency}
    \end{small}
  \end{center}
The ADD is typically between 1.8 and 3.6 over a range of different languages \citep{liu2008dependency}.

\subsection{Unlabeled Attachment Score}
Accurate computation of the Average Dependency Distance (ADD) necessitates starting with a correct dependency tree. Therefore, evaluating how well various models perform dependency parsing is crucial. The quality of a dependency parser can be assessed by calculating the average \emph{Unlabeled Attachment Score (UAS)} over a set of sentences. The UAS represents the proportion of dependency relations (i.e., the connections between words) that the parser correctly identifies when compared to the gold-standard (correct) tree, disregarding the dependency labels (no labels are shown in the tree above). In the example above, if a model would predict an arrow from ``made'' to ``fridge'', rather than the correct arrow from ``pie'' to ``fridge'', the UAS would be 6/7, or about 0.86.

\subsection{Swedish readability index LIX}
{\em LIX}, a readability index suggested by \citet{bjornsson1968lasbarhet}, is computed as $A/B + 100C/A$, where $A$ is the number of words in the text, $B$ the number of sentences, and $C$ is the number of words longer than six letters. Higher LIX values indicate a more advanced text: Typically LIX$<$30 is considered an easy text, whereas LIX$>$50 is advanced, and LIX$>$60 very advanced (e.g., research papers).


\section{Method}
We randomly selected 345 university-level and high-school essays\footnote{From {\tt https://www.diva-portal.org} and {\tt https://www.mimersbrunn.se/}, respectively.}.  All essays were written in Swedish before 2018 (to ensure that the author had not used any generative AI writing tool). From each essay, we randomly selected one paragraph (averaging 71 $\pm$ 15 tokens) for the LIX calculations, and one sentence (averaging 26 $\pm$ 8 tokens) for the dependency parsing experiment. Examples of paragraphs and sentences can be found in the Appendix. 

We computed the ground-truth LIX values by a Python script implementing the LIX formula. The ground truth dependency trees were produced using the Stanza library\footnote{{\tt https://stanfordnlp.github.io/stanza/}}. 
We then asked the four models to compute the LIX score of each paragraph, and then to analyze our selected sentences and print their dependency trees. We instructed the models to print one word per row, on the following format:
\begin{quote}
{\tt 1, Han, 2, 1}
\end{quote}
i.e., the word index, the word itself, the index of the headword, and the dependency distance between the word and its head. The exact prompts can be found in the appendix. Finally, we asked each model to compute the average dependency distance for each sentence.

\section{Results}

\subsection{Lix}

The LIX values calculated by the models highlight a clear trend in their ability to accurately compute readability scores. The errors in LIX values range from 7.4 to 20.9, with the best performance achieved by o1-mini, which recorded the lowest LIX error of 7.4. In contrast, llama-70b had the highest error at 20.9, indicating room for improvement in its computation of readability metrics.

Intermediate performance was observed with models like Gemini-2.0-flash and GPT-4o-mini, which achieved LIX errors of 10.42 and 9.2, respectively. These results suggest that newer and more advanced models tend to perform better in calculating LIX, aligning with improvements in their general capabilities.

The consistency of the results across models also indicates that LIX computation may serve as a valuable metric for assessing a model’s ability to process linguistic features effectively. This evaluation is particularly relevant for applications in natural language understanding where readability and text complexity play crucial roles.


\subsection{Correlation between MMLU and LIX error}

To evaluate the relationship between the performance of large language models (LLMs) on the Massive Multitask Language Understanding (MMLU) benchmark and their linguistic complexity, as measured by the LIX Error, a Pearson correlation coefficient ($r$) was calculated.

The MMLU score represents the accuracy of the models on a standardized benchmark assessing general knowledge and reasoning abilities across multiple domains, while the LIX Error reflects the linguistic complexity or readability of the text generated by the models.

The Pearson correlation coefficient measures the linear relationship between these two variables, where $r$ values range from $-1$ (perfect negative correlation) to $+1$ (perfect positive correlation), with $0$ indicating no linear relationship.

The analysis was conducted using Python's \texttt{scipy.stats.pearsonr} function, which computes both the correlation coefficient ($r$) and the corresponding two-tailed $p$-value. The $p$-value was used to assess the statistical significance of the observed correlation, with a threshold of $\alpha = 0.05$ applied to determine significance.

The computed correlation coefficient was $r = -0.875$, indicating a very strong negative linear relationship between MMLU scores and LIX Errors. Furthermore, the $p$-value of $0.026$ suggests that this relationship is statistically significant, demonstrating that as models achieve higher MMLU scores, their LIX Errors consistently decrease, reflecting improved accuracy in linguistic complexity computations.

\begin{table}[h]
\centering
\begin{tabular}{|l|c|c|}
\hline
\textbf{Model} & \textbf{MMLU} & \textbf{LIX Error} \\
\hline
Gemini-1.5-pro & 85.9 & 19.72 \\
Gemini-2.0-flash & 87.0 & 10.42 \\
llama-70b  & 86.0 & 20.9         \\
llama-70b 3.3 & 86.0 & 18.64 \\

GPT-4o-mini  & 88.7                & 9.2          \\
o1-mini  & 90.8                & 7.4           \\
\hline
\end{tabular}
\caption{MMLU and LIX error for various models}
\label{table:MMLU_LIX}
\end{table}

\subsection{Dependency parsing}
All the models have a grasp of the fundamentals of dependency parsing: the words in the sentence are indexed from 1 up to the number of words, and the index of the head of any word (almost) always is a number in this range. To accurately measure the average UAS, however, one cannot simply go through all the words in a sentence and check that the model's predicted head word index matches that of the ground truth. This is because the model might tokenize the sentence differently from what is expected in the ground truth dependency tree, which would result in the words being numbered differently from the ground truth tree. As an example, we noticed that quoted words, like ``hårda'' (Eng.\ ``{\em hard}'') was tokenized as three tokens ({\tt ", hårda, "}) by o1-mini, which is arguably more correct than the ground truth tokenization, which considered {\tt "hårda"} to be a single token. 

Instead of an index-based comparison, we carefully checked the dependency relations based on the words themselves to see if the model's predictions matched the ground truth. The results are shown in Table 2. We also calculated the Pearson correlation coefficient between MMLU and Add diff 1 to be 
-0.519, with a two-tailed p-value of 0.370. This indicates a moderate negative correlation that is not statistically significant.

\subsection{Average dependency distance}

\begin{table}[h]
\begin{center}
    \begin{tabular}{|l|c|c|}
        \hline
        {\bf Model} & {\bf ADD diff 1} & {\bf ADD diff 2}\\
        \hline
        Gemini-1.5-pro & 1.02 & 3.54\\
        Gemini-2.0-flash & 0.66 & 0.41\\
        llama-70b & 0.88 & 0.64\\
        GPT-4o-mini & 0.97 & 1.38\\
        o1-mini & 0.64 & 0.12\\      
        \hline
    \end{tabular}
    \caption{ADD diff 1 = Average absolute difference between ADD in the ground truth dependency trees and the dependency trees produced by various models. ADD diff 2 = Average absolute difference between the actual ADD in the dependency trees computed by the model, and the ADD reported by the models themselves.}
    \label{table:add}
    \end{center}
   \end{table}
   Table \ref{table:add} shows the average absolute difference between the ADD computed from the ground truth dependency trees and the ADD computed from the dependency trees produced by the various models (``ADD diff 1''). o1-mini is best in this category as well. In addition, we can note that the difference between the ADD values reported by o1-mini and the actual ADD values of its produced trees (``ADD diff 2'')  is very small. That is, it is not necessary to use some external program to calculate the ADD of a sentence based on a dependency tree constructed by the model, but one can directly ask the model for the ADD value and expect to get a quite accurate result. Note that this is not the case for the other models. Both ADD diff~1 and ADD diff 2 are negatively correlated with MMLU scores (-0.83 and -0.63, respectively).


Table 3 (in the appendix) shows that, quantitatively, o1-mini is the best-performing model of the five, though the quality of its inferences vary between different parts-of-speech. For example, o1-mini is good at inferring the head for adjectives and determiners (whose head is most often the noun that follows closely afterwards in the text), but bad at placing coordinating conjunctions correctly. A common example in the latter category are noun phrases consisting of two nouns with a conjunction in between, e.g.\ ``organisation och orientering'' (Eng.: ``{\em organisation and orientation''}). The correct structure according to the Universal Dependencies guidelines (and Stanza) is:
  
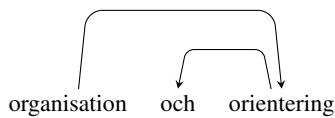
\begin{center}
    \begin{small}
    \begin{dependency}[hide label]
      \begin{deptext}[column sep=0.3cm]
        organisation \& och \& orientering \\
      \end{deptext}
      \depedge{1}{3}{}
      \depedge{3}{2}{}
    \end{dependency}
    \end{small}
  \end{center}
i.e., the first word of the conjunction is considered to be head of the whole phrase. However, no model is consistent in this regard: sometimes the first noun is considered the head, sometimes the second, and sometimes the conjunction (but arguably that this choice is somewhat arbitrary anyway).

Another phenomenon that all models get (consistently) wrong is the root node in the presence of a predicative, e.g., ``Varje project är unikt'' (Eng. ``{\em Every project is unique}''). Though normally the main verb is the root word in each sentence, this is not the case when the main verb is a copula (the verb ``be''). Instead, the predicative ``unique'' should be the root word. Every model got this wrong, indicating instead that ``är'' (``{\em is}'') should be the root.

o1-mini predicted the role of punctuation signs in the sentence much better than the other models. Gemini, in particular, often skipped the punctuation signs altogether (although it was explicitly instructed to consider all punctuation signs as tokens), leading to a very poor result in this category. 

\section{Discussion}
Understanding the inner workings of LLMs and evaluating their performance are tasks that hopefully can be done in parallel. In this work exploring language complexity and LLMs we showed that language complexity evaluation can be used to evaluate overall model performance. As these models become ubiquitous, quick methods for evaluation and understanding become more important. The strengths of our technique is its simplicity and language independence. The fact that we do not evaluate the content of language and instead evaluate the understanding of the structure of language makes our technique a reasonable candidate for evaluation of reasoning in LLMs. In the same way working memory tests are a noisy proxy for intelligence in humans, our tests can be seen as a working memory test for LLMs where models with better "working memory"; reasoning models outperform other models.

\subsection{Conclusion}
Language complexity metrics like LIX and ADD provide valuable insights into LLM capabilities. Despite advances in fluency, LLMs remain prone to errors in explicit calculations and syntactic fidelity. These findings suggest that complexity metrics can efficiently complement broader evaluation benchmarks.

\section{Limitations}
Although our findings demonstrate that language complexity measures (e.g., LIX) can serve as a useful—albeit noisy—zero-shot proxy for general LLM abilities, several limitations merit discussion:

\paragraph{Single run:}
Every LIX-computation and dependency-parsing was only done once per sentence and model.

\paragraph{Scope of Evaluation:}
First, our experiments focus exclusively on Swedish texts drawn from university and high-school essays. These sources may not capture the full breadth of linguistic styles and complexities found in other domains, genres, or languages. Consequently, our results may not generalize to languages with different morphological structures or to more specialized text types such as technical documents or colloquial dialogues.

\paragraph{Reliance on Proprietary Models:}
We primarily evaluated closed-source, proprietary models whose internal architectures, tokenization schemes, and training data are neither publicly accessible nor standardized.

\paragraph{Tokenization and Computational Fidelity:}
Our analysis relies on each model’s tokenization process, which can differ substantially across architectures and affect the accuracy of LIX or ADD computations. Because we depend on the models to generate the appropriate tokens and numerical outputs, discrepancies in tokenization can introduce systematic errors in measured performance. 

\paragraph{Rapid Model Evolution:}
Large Language Models evolve rapidly through fine-tuning and architectural enhancements. Performance measured at a particular snapshot in time may not generalize to newer versions of the same model. Ongoing and systematic evaluations are therefore essential to capture how fast-changing model iterations handle language complexity tasks.

\paragraph{Data Sampling:}
Finally, we used a limited number of paragraphs and sentences for LIX and dependency parsing evaluations. While this was sufficient to identify consistent errors and variance across models, larger datasets—potentially spanning multiple domains—would improve robustness and strengthen the statistical reliability of our conclusions.

\bibliography{custom}

\appendix

\section{Appendix}
\label{sec:appendix}

\begin{table*}[t!]
\label{addtable}
\centering
\begin{tabular}{|l|c|c|c|c|c|}
\hline
\textbf{Label / Model} & \textbf{Gemini-pro} & \textbf{Gemini-2.0-flash} & \textbf{llama-70b} & \textbf{GPT-4o-mini} & \textbf{o1-mini} \\ 
\hline
ADJ & 0.49 & 0.57 & 0.56 & 0.59 & 0.71 \\
ADP & 0.23 & 0.10 & 0.28 & 0.22 & 0.13 \\
ADV & 0.43 & 0.56 & 0.47 & 0.47 & 0.62 \\
AUX & 0.21 & 0.23 & 0.23 & 0.16 & 0.32 \\
CCONJ & 0.18 & 0.16 & 0.23 & 0.18 & 0.08 \\
DET & 0.38 & 0.62 & 0.46 & 0.47 & 0.76 \\
NOUN & 0.31 & 0.30 & 0.34 & 0.31 & 0.37 \\
NUM & 0.27 & 0.35 & 0.28 & 0.35 & 0.45 \\
PART & 0.33 & 0.42 & 0.40 & 0.26 & 0.49 \\
PRON & 0.30 & 0.32 & 0.39 & 0.33 & 0.46 \\
PROPN & 0.29 & 0.19 & 0.27 & 0.28 & 0.30 \\
PUNCT & 0.04 & 0.28 & 0.16 & 0.19 & 0.49 \\
SCONJ & 0.10 & 0.33 & 0.19 & 0.07 & 0.18 \\
VERB & 0.16 & 0.28 & 0.22 & 0.19 & 0.26 \\
\hline
Micro-average (UAS) & 0.28 & 0.32 & 0.33 & 0.30 & 0.38 \\
Macro-average (UAS) & 0.27 & 0.34 & 0.32 & 0.29 & 0.40 \\
\hline
\end{tabular}
\caption{The table shows how well various models predicted the correct heads to words of various parts-of-speech, relative to the ground truth as given by Stanza. The final rows shows the UAS averaged over all datapoints (micro-average) and over all part-of-speech classes (macro-average). We are using the standard set of parts-of-speech according to Universal Dependencies ({\tt https://universaldependencies.org/u/pos/index.html})}.
\end{table*}

\subsection{Prompts used}

\begin{tcolorbox}[title=Average Dependency Distance (ADD)]
I would like you to print the dependency parsing result for a given Swedish sentence. Print the result with one word on each row, on the following form: 

\begin{tabular}{l}
1, Han, 2, 1\\
2, köper, 0, 0\\
3, en, 4, 1\\ 
4, bok, 2, 2
\end{tabular}

where the first number is the word index, the second column is the word itself, the third column is the index of the head word, and the last number is the dependency distance (i.e. the absolute difference between the index and the head word index). The root word should have head word=0 with a distance of 0. Finally, print the average of all the dependency distances in the sentence. Here is the sentence: "{\em text}"

\end{tcolorbox}

\begin{tcolorbox}[title=Readability index (LIX)]
Analyze the following Swedish text and compute a LIX readability score. Here's how to calculate LIX:

1. Calculate the average sentence length (number of words / number of sentences)
2. Calculate the percentage of long words (words with more than 6 characters)
3. Add these two numbers together to get the LIX score

Important: LIX scores typically range from 20 to 60, with:
- 20-30 being very easy text
- 30-40 being easy text  
- 40-50 being medium difficulty
- 50-60 being difficult text
Any score above 60 is extremely rare and likely indicates an error in calculation.

Please show your calculations and provide the final score in the format 'LIX=' followed by the number.

Here is the text to analyze:  $\langle$text$\rangle$
\end{tcolorbox}




\subsection{Example texts used for the LIX computations}

\subsection*{u1}
Finns det någon mening med att studera histora?Jag anser att det gör det.  Som ett argument för det kan man använda det klassiska uttrycket:	- Man lär sig av sina misstag. Och det gör man.  Misstag som man gjort ligger bakom en och det som ligger bakom en är också historia.  ven om det rör sig om att man som 2-ring lär sig att inte springa in ett träd för att man får väldigt ont då.

\subsection*{u2}
En dator brukar delas in i tre olika delar.  Nämn dessa delar och förklara varför man valt just denna indelning. Datorn brukar som sagt delas in i tre olika delar.  En centralenhet, en indataenhet och en utdataenhet.  Indataenheter är som det låter, saker som vi använder för att skicka in data till datorn, tangentbordet är ett bra exempel, scanner och gamepads är två andra exempel.

\subsection*{u3}
Pizzans historia börjar antagligen såhär:"Det var mycket strider förr.  Folk reste omkring, men tallrikarna blev smutsiga, och diska dem tog för lång tid.  Men så var det nån smart hjärna som kom på att man kunde göra tallrikarna av bröd! Ja, dom åt på tallrikarna av bröd, men efter måltiden slängde dom brödet.  Långt senare började en resturang, kallad som Bruno vid berget Vezuvio, att lägga tomater, mozarella och persilja på, eller om det var nån annan krydda.  Men den var endast känd där, i byn.

\subsection*{u4}
Historien om WUFC handlar om ett av de största graffiticrewen i Stockholm.  Journalisten Björn Almqvist har följt några av graffitimålarna som är med i WUFC under flera års tid och fotat alla tunnelbanor som de har målat på, alla väggar de har målat på och på deras resor runt om i världen.  Min personbeskrivning på graffitimålaren QUE skulle vara att han håller på med graffiti för att uttrycka sig själv och sina känslor.  Que är inte den som tar till våld i första taget utan försöker att lösa sina konflikter med ord.  Det är lite svårt att göra en miljöskrivning ur denna bok, men om jag skulle vilja beskriva någon plats skulle det vara hemma hos Que.

\subsection*{u5}
Jag valde att skriva om när vallonerna emigrerade hit från Vallonien till Sverige.  Dels valde jag det för att de har spelat en stor roll i vår svenska smidesindustri, dels för att jag fick höra att vi hade vallonblod i släkten och tyckte därför att det var ett intressant ämne. 

Vallonerna emigrerade hit från Vallonien, ett område som ligger mellan Belgien och Frankrike, under 1600-talet och en bit framåt.  När jag gick över statistik om emigration från Vallonien fann jag att Sverige var bland de fem regioner dit vallonerna flyttat till mest, tillsammans med Flandern, Brasilien, Argentina och USA (Wisconsin, framförallt).  Man kan då fråga sig vad orsakerna var till att Sverige var så lockande för valloner och varför de emigrerade just hit.

\subsection*{h1}
Under 1800-talets slut bodde svenskarna på landet och det var bara två av tio som bodde i städerna.  Vid år 2000 bodde det så mycket som nio av tio av svenskarna i städer och tätorter.  Urbaniseringen, som detta kallas, har gjort så att hälften av dagens befolkning bor i de femton största städerna.  Urbaniseringen har gjort många förändringar när det gäller bosättningen i landet under dessa år.  Just nu bor cirka 85 \% av befolkningen i städerna och i tätorter.

\subsection*{h2}
FN bildades den 24 oktober år 1945.  Deras föregångare var Nationernas förbund, men dom lyckades inte så bra.  Nationernas förbund klarade inte va pressen efter första världskriget.  Efter 2: a världskriget så bestämde sig 51 länder att bilda FN, Förenta nationerna.  I början av 1994 så var de 184 länder som var medlemmar i FN.

\subsection*{h3}
I Nigeria varierar klimatet väldigt mycket beroende på var i landet man befinner sig.  I söder alltså där Nigeria har kontakt med Atlanten är det varmt året om, ungefär 25 grader.  Där regnar det också väldigt mycket ungefär 2000-3000mm per år.  Anledningen att det regnar så mycket där är att Nigerias syd kust ligger så nära Atlanten.  På morgonen och lite in på dagen avdunstar vattnet ifrån Atlanten som senare regnar ned på eftermiddagen.

\subsection*{h4}
Jag skulle vilja säga att imperialismen började så långt tillbaks som för2000år sedan, och då tänker jag främst på rommarna som hade erövrat stora delar Europa och även delar av Asien.  Rommarna styrde dem kända världen som dem mest överlägsnaste ledaren i världen.  Jag skulle även vilja kalla spanjorerna och portugiserna för imperialister.  De tog över helavärldsdelar och tog dem som kolonier.  Dessa länder ihop med tyskland ochEngland koloniserade även hela Afrika och i och med det kom också handeln med slavar som skeppades i massor till "den nya världen" Amerika.

\subsection*{h5}
I Sverige har vi en av världens bästa lagar mot diskriminering inom arbetslivet.  Det skall inte spela någon roll varifrån man kommer eller vilken hudfärg man har, det skall vara den mest kvalificerade på jobbet.  Allting låter bra så här långt.  Hur kommer det sig då att Sverige har ett yrke som består av tandläkare, läkare, ingenjörer och andra högutbildade? Det är inte någon ny specialutbildning för medicinstuderande utan taxichaufförer.  Hur kommer detta sig?SituationMånga av de invandrare som kommer till Sverige har inte varit några outbildade bidragstagare i sitt hemland.
\label{sec:appendix}

\subsection{Example sentences used for dependency parsing}
\begin{enumerate}
    \item Även om det rör sig om att man som 2-åring lär sig att inte springa in ett träd för att man får väldigt ont då.
    \item Indataenheter är som det låter, saker som vi använder för att skicka in data till datorn, tangentbordet är ett bra exempel, scanner och gamepads är två andra exempel.
    \item Långt senare började en resturang, kallad som Bruno vid berget Vezuvio, att lägga tomater, mozarella och persilja på, eller om det var nån annan krydda.
    \item Journalisten Björn Almqvist har följt några av graffitimålarna som är med i WUFC under flera års tid och fotat alla tunnelbanor som de har målat på, alla väggar de har målat på och på deras resor runt om i världen.
    \item 
    När jag gick över statistik om emigration från Vallonien fann jag att Sverige var bland de fem regioner dit vallonerna flyttat till mest, tillsammans med Flandern, Brasilien, Argentina och USA (Wisconsin, framförallt).
    \item 
    Urbaniseringen, som detta kallas, har gjort så att hälften av dagens befolkning bor i de femton största städerna.
    \item 
    Efter 2: a världskriget så bestämde sig 51 länder att bilda FN, Förenta nationerna.
    \item 
    På morgonen och lite in på dagen avdunstar vattnet ifrån Atlanten som senare regnar ned på eftermiddagen.
    \item 
    Jag skulle vilja säga att imperialismen började så långt tillbaks som för2000år sedan, och då tänker jag främst på rommarna som hade erövrat stora delar Europa och även delar av Asien.
    \item 
    Många av de invandrare som kommer till Sverige har inte varit några outbildade bidragstagare i sitt hemland.
\end{enumerate}

\end{document}